\documentclass[11pt]{article}
\usepackage[dvips]{graphicx}
\newcommand{\kb}{{knowledge}}
\newcommand{\colw}{{CoLweb}}

\newcommand{\oo}{\bot}            
\newcommand{\pp}{\top}            

\newcommand{\mla}{\mbox{{\Large $\wedge$}}}

\newcommand{\pst}{\mbox{\raisebox{-0.01cm}{\scriptsize $\wedge$}\hspace{-4pt}\raisebox{0.16cm}{\tiny $\mid$}\hspace{2pt}}}
\newcommand{\gneg}{\neg}                  
\newcommand{\mli}{\rightarrow}                     
\newcommand{\cla}{\mbox{\large $\forall$}}      
\newcommand{\cle}{\mbox{\large $\exists$}}        
\newcommand{\mld}{\vee}    
\newcommand{\mlc}{\wedge}  
\newcommand{\ade}{\mbox{\Large $\sqcup$}}      
\newcommand{\ada}{\mbox{\Large $\sqcap$}}      
\newcommand{\add}{\sqcup}                      
\newcommand{\adc}{\sqcap}                      

\newcommand{\tlg}{\bot}               
\newcommand{\twg}{\top}               

\newtheorem{theoremm}{Theorem}[section]
\newtheorem{conditionss}{Condition}[section]

\newtheorem{definitionn}[theoremm]{Definition}
\newtheorem{lemmaa}[theoremm]{Lemma}
\newtheorem{notationn}[theoremm]{Notation}
\newtheorem{propositionn}[theoremm]{Proposition}
\newtheorem{conventionn}[theoremm]{Convention}
\newtheorem{examplee}[theoremm]{Example}
\newtheorem{remarkk}[theoremm]{Remark}
\newtheorem{factt}[theoremm]{Fact}
\newtheorem{exercisee}[theoremm]{Exercise}
\newtheorem{questionn}[theoremm]{Open Problem}
\newtheorem{conjecturee}[theoremm]{Conjecture}

\begin{document}
\begin{center}
 {\Large Implementing Dynamic Programming in Computability Logic Web} \\[15pt]
 Keehang Kwon \\
Dept. of Computer Engineering, DongA University \\
khkwon@dau.ac.kr
\end{center}

We  present a novel  definition of an algorithm and its corresponding algorithm language called \colw.
  The merit of \colw\cite{Jap03}  is that it makes algorithm design so versatile.
   That is, it forces us to a high-level, proof-carrying, distributed-style approach to
   algorithm design for 
   both  non-distributed computing and distributed one.
 We argue that this  approach
  simplifies algorithm design. In addition, it unifies other approaches including
 recursive logical/functional algorithms, imperative algorithms, object-oriented imperative algorithms, neural-nets, interaction nets, proof-carrying code, etc.

As an application, we  refine  Horn clause definitions into two kinds:
blind-univerally-quantified (BUQ) ones and parallel-universally-quantified (PUQ) ones. BUQ definitions
corresponds to the
traditional ones such as those in Prolog where knowledgebase is $not$ expanding and its
proof procedure is based on the backward chaining. 
On the other hand, in PUQ definitions, knowledgebase is $expanding$ and its proof procedure leads to 
forward chaining and {\it automatic memoization}.

\section{\colw\ as a unifying framework}\label{sec:intro}

Computer science lacks a unifying computing model. It consists of diverse
models such as pseudocode, petri-net, interaction net, web languages, etc.
Computability-logic web (\colw) is a recent attempt to provide a unifying
computing model for distributed computing with the following
principle:

\[ computation\ as\ game\ playing \]
\noindent 
In particular, it integrates all the concepts widely used in AI as well as in
everyday life -- strategies, tactics, proofs, interactions, etc. 

In this paper, we slightly extends \colw\ with class agents.
 A {\it class} agent is a cluster of agents in a compressed form.
It is  useful for applications such as dynamic programming.

We show that \colw, if scaled down,  is also a candidate unifying model for a 
single, non-distributed computing. That is, it is an appealing alternative to pseudocode. 
This is not so surprising, because  a block of memory cell within 
a single machine can be seen as an (local) agent.  From this view,
a machine with local memories can be
seen as a  multi-agent system where each agent shares a CPU.


\section{Implementing Dynamic Programming}\label{sec:intro}

In the traditional  logic programming approaches such as Prolog,   their knowledgebase is
$not$ expanding and their underlying proof procedure is based
on the backward chaining such as uniform proof. 
For example, a  $fib$ function is  specified as: \\

$   fib(0) = 1 $

$   fib(1) = 1$

$  \forall x,y,z [ (fib(x,y) \mlc  fib(x\!+\!1,z)) \mli fib(x\!+\!2,y+z)]. $ \\

Note that the third definition --  called  {\it blind-universally-quantified
definition}  in \cite{Jap03} -- is true independent of $x$.
For example, to compute $\ade xfib(3,x)$,  the machine simply returns 3 without expanding the BUQ-definiiton.

However, this procedure  discards lemmas which are generated as new instances of
the existing knowledgebase. Often these lemmas are useful and discarding them often leads to clumsy code when
it deals with dynamic programming.

There have been different approaches to  this topic such as tabled logic programming. These approaches, however,
lead to slow performance. In this paper, we introduce a novel  approach  which is based on PUQs.
As we shall see from the semantics of $\mla$,
  its proof mechanism  is required to add lemmas to the program, rather than discards them.
 These lemmas play a role similar to automatic memoization.

For example, a  $fib$ relation is  specified as: \\

$   fib(0,1) $

$   fib(1,1)$

$ \mla x,y,z\ [(fib(x,y) \mlc  fib(x\!+\!1,z) \mli fib(x\!+\!2,y+z)]$ \\

Note that the third definition -- which we call {\it PUQ
definition}  -- is compressed
and  needs to be $expanded$ during execution. 
For example, to compute $fib(3,3)$, our proof procedure adds the following: \\

$ fib(0,1)\mlc fib(1,1) \mli     fib(2,2)$

$ fib(1,1)\mlc fib(2,2) \mli fib(3,3). $ \\

One consequence of our approach is that it supports dynamic programming in  a clean code.

\section{\colw\  Algorithms}\label{sec:logic}

Although algorithm is one of the central subjects, there have been
little common understandings of what an algorithm is.
We believe the following definition is a simple yet
 quite compelling definition:  \\

 An  algorithm for an agent $d$  who can perform a task $T$, written as
 $d=T$, is  defined recursively as follows: \\ \\

 Algorithm $alg(d = T)$ \\

 Step 1: Identify a set of agents  
     $c_1 =  T_1$, \ldots, $c_n = T_n$ such that they can collectively perform $T$. \\

 Step 2. Call recursively each of the following:  $alg(c_1=T_1),\ldots, alg(c_n=T_n)$. \\

\noindent Note that Step 2 is missing from the traditional definition of algorithms, i.e., 
algorithm as a sequence of instructions to perform a task. We view instructions something that are not fixed and
Step 2 is added for  manipulating the instructions.  It turns out that Step 2 is the key idea which
makes algorithm design so  interesting, compared to  the traditional definition.  In other words, our approach
corresponds to $n$-level-deep algorithms, while the traditional one to 1-level-deep.

In the above, $c = T$ represents  an agent $c$ who can do task $T$.
In the traditional developments of declarative algorithms, $T$ is limited
to simple tasks such as  recursive functions or Horn clauses. 
Most complex
tasks such as interactive ones  are not permitted. 
In our algorithm design, however,  interactive tasks are allowed.

To define the class of computable tasks, we need a specification language. 
 An ideal  language would support an optimal
 translation of the tasks. Computability logic(CoL)\cite{Jap03} is exactly what we need here.

\section{Preliminaries}\label{s2}

In this section a  brief overview of CoL is given. 

There are two players: the machine $\pp$ and the environment $\oo$.

There are two sorts of atoms: {\em elementary} atoms $p$, $q$, \ldots to represent elementary games, and {\em general atoms} $P$, $Q$, \ldots to represent any, not-necessarily-elementary, games. 

\begin{description}
\item[Constant elementary games]  $\twg$ is always a true proposition, and $\tlg$ is always a false proposition.

\item[Negation]
 $\gneg$ is a role-switch operation: For example, $\gneg (0=1)$ is true,
while $(0=1)$ is false.

\item[Choice operations]
The choice group of operations:  $\adc$, $\add$, $\cla$ and $\cle$ are defined below.

$\ada xA(x)$ is the game where, in the initial position, only $\oo$ has a legal move which consists in 
choosing a value for $x$. After $\oo$ makes a move $c\in\{0,1,\ldots\}$, 
the game continues as $A(c)$. 
$\cla xA(x)$ is similar, 
only here the value of $x$ is invisible. $\ade$ and $\cle$ are 
symmetric to $\ada$ and $\cla$, with
the  difference that now it is $\pp$ who makes an initial move.

\item[Parallel operations]
Playing $A_1\mlc\ldots\mlc A_n$ means playing the $n$ games concurrently.  In order to win,  $\pp$ needs to win in each of $n$ games. Playing  $A_1\mld\ldots\mld A_n$ also means playing the $n$ games concurrently.  In order to win,  $\pp$ needs to win  one of the games. To indicate that a given move is made in the $i$th component, the player should prefix it with the string ``$i.$".  
The operations $\pst A$ means an infinite parallel
game $A\mlc\ldots\mlc A\mlc\ldots$.
 To indicate that a given move is made in the $i (i>1)$th component, we assume the player should 
 first replicate $A$ and then prefix it with the string ``$i.$". 
 
 In addition,  we use the following notation:
 
 \[ \mla x_m^n A =_{def}  A(m/x) \mlc \ldots \mlc A(n/x). \]
 
\item[Reduction]
 $A\mli B$ is defined  by $\gneg A\mld B$.
Intuitively, $A\mli B$ is the problem of reducing $B$ ({\em consequent}) to $A$ ({\em antecedent}).  

\end{description}

\section{\colw }\label{sec:intro}

\colw\ is a multi-agent version of computability logic. It consists of a set of agent declarations. An example
looks like the following  form: \\\\

$\alpha_1 = F_1$ \\

$\alpha_2 = F_2$ \\

$\alpha_3 = F_3^{\alpha_1} \mlc F_4^{\alpha_1,\alpha_2}$ \\

$\alpha_4 = \alpha_2^{\alpha_1}$ \\

$\mla x\ \alpha(x) = F(x)$ \\

Here $\alpha_1$ is an agent name and $F_1$ is called the \kb\ of $\alpha_1$.
Similarly for $F_2$ is the \kb\ of $\alpha_2$.

The agent $\alpha_3$ has $F_3$ and $F_4$ as its \kb,  provided that $F_3$ is a
logical consequence of $\alpha_1$ and $F_4$ a logical consequence of
$\alpha_1, \alpha_2$ combined.
If $F_3$ is not
logical consequence of $\alpha_1$,  the above will be converted to: \\

\[ \alpha_3 = \pp \mlc F_4^{\alpha_1,\alpha_2}. \]

In $\alpha_4$, $\alpha_2$ is a macro call, meaning that 

\[ \alpha_4 = F_2^{\alpha_1}. \]

The agent $\mla x \alpha(x) = F(x)$ is a {\it class agent} which 
means the following: \\

\[ (\alpha(1) = F(1))  \mlc (\alpha(2) = F(2)) \mlc \ldots \mlc \ldots \]

As an example, consider the following  \colw: 

\[ /m(0) = q \]
\[ \mla x\ /m(x') = p \land /m(x) \]

\noindent Given this definition, $p \land (p \land (p \land q)))$ can be represented simply as $ /m(0''')$.  We assume in the above that
$'$ is the number-successor function.

\section{Example}

\newenvironment{exmple}{
 \begingroup \begin{tabbing} \hspace{2em}\= \hspace{3em}\= \hspace{3em}\=
\hspace{3em}\= \hspace{3em}\= \hspace{3em}\= \kill}{
 \end{tabbing}\endgroup}
\newenvironment{example2}{
 \begingroup \begin{tabbing} \hspace{8em}\= \hspace{2em}\= \hspace{2em}\=
\hspace{10em}\= \hspace{2em}\= \hspace{2em}\= \hspace{2em}\= \kill}{
 \end{tabbing}\endgroup}

Recursive algorithms focuses on designing algorithms with only one agent. On the contrary,
in the  multi-agent (or object-oriented) approach, there is no limit regarding the number of agents involved in 
developing an algorithm.
Therefore, the main virtue of the multi-agent approach 
is that much more diverse  algorithms are possible.

 To support multi-agent programming,
computability logic  allows the introduction of a {\it location/agent/object}
for knowledgebases. We called this computability logic web (CoLweb) in our
previous works. For example, assume that
$fib(X,Y)$ is stored at a location $/a[X]$ and the rule

\[ \cla y\cla z (fib(X,y) \mlc  fib(X\!+\!1,z)) \mli fib(X\!+\!2,y\!+\!z) \]
\noindent is stored at $/b[X+2]$.
  Then $fib$ can be
rewritten as: \\

\noindent  $\mla x_0^{\infty}\ /b[x\!+\!2] =    \cla y\cla z [(fib(x,y) \mlc  fib(x\!+\!1,z)) \mli fib(x\!+\!2,y\!+\!z)]$ \\
  
\noindent  $/a[1] = fib(1,1)$ \\

\noindent   $/a[2] = fib(2,1)$ \\

\noindent  $\mla x_0^{\infty}\ /a[x\!+\!2] =    (\ade y  fib(x+2,y))^{/a[x],/a[x\!+\!1],b[x+2]}  $ \\

\noindent   $/fib = \ada n (\ade y  fib(n,y)^{/a[n]})$ \\

\noindent   $/query = \ade y  fib(4,y)^{/fib}$ \\

\noindent
Now consider an expression $/query$. 
This expression  tries to solve $\ade y fib(4,y)$ relative to $/fib$.
It then executes $fib$ by trying to solve $\ade y fib(4,y)$ relative to $/a[4]$. 
It eventually creates the following which is a form of memoization: \\

$ /a[3] =  fib(3,2)$

$  /a[4] =  fib(4,3)$ 

  $\mla x_3^{\infty}\ /a[x\!+\!2] = (\ade y  fib(x+2,y))^{/a[x],/a[x\!+\!1],b[x+2]}  $ \\
\hspace*{12em} 

\noindent
Similarly  for the class agent $/b$. Note also that
the above code is nothing but an object-oriented
programming in a distilled form. That is, in object-oriented terms, 
$/a[1],/a[2]$ are regular objects and $/a[x\!+\!2],/b[x\!+\!2]$ are {\it class}
objects.  
  
\section{Some Variation}

The Fibonacci implementation in the previous section is quite complex and
is hard to prove its correctness. We will present a simpler version. \\

\noindent  $\ /b =     \cla x \cla y\cla z (fib(x,y) \mlc  fib(x\!+\!1,z)) \mli fib(x\!+\!2,y\!+\!z))$ \\
  
\noindent  $/a[1] = fib(1,1)$ \\

\noindent   $/a[2] = fib(2,1)$

\noindent  $\mla x_0^{\infty}\ /a[x\!+\!2] =    (\ade y (fib(x+2,y))^{/a[x],/a[x\!+\!1],/b}  $ \\

\noindent   $/fib = \ada n ((\ade y  fib(n,y))^{/a[n]})$ \\
\noindent   $/query = (\ade y  fib(4,y))^{/fib}$ \\

The above is somewhat different from the old version in that the agent $b$ is not a class agent
anymore. The agent eliminates $\mla$ in favor of $\cla$-quantifier.
One consequence is that $b$ is not expanding anymore.

\noindent
Now consider an expression $/query$. 
This expression  tries to solve $\ade y fib(4,y)$ relative to $/fib$.
It then executes $fib$ by trying to solve $\ade y fib(4,y)$ relative to $/a[4]$. 
It eventually creates the following which is a form of memoization: \\

$ /a[3] =  fib(3,2)$ 

$  /a[4] =  fib(4,3)$ 

  $\mla x_3^{\infty}\ /a[x\!+\!2] = (\ade y (fib(x\!+\!2,y))^{/a[x],/a[x\!+\!1],/b}$ \\
  \noindent

\section{ Algorithm Design via CoLweb}

There are  traditional  approaches to algorithm design (for small-size problems) and software design (for big-
size ones):

\[ {\rm algorithm\ design\ =\ pseudocode,  etc} \]

\[ {\rm software\ design\ =\ UML} \]

\noindent We propose an alternative approach:

\[ {\rm algorithm\ design\ =\ CoLweb} \]

\[ {\rm software\ design\ =\ CoLweb} \]

We compare some approaches.

\begin{enumerate}

\item  CoLweb algorithms 

\item imperative algorithms

\item object-oriented imperative algorithms

\item logical/recursive algorithms

\end{enumerate}

(1) is a high-level, lemma-based approach to algorithm design and 
has the following features: (a)  consists of a sequence of $big$ steps 
(b) execution from bottom-up, (c) support parallelism, and (d) support proof-carrying codes.
We consider this approach the best, leading to  clean, versatile  yet efficient codes. 

Both (2),(3) are  inferior versions of (1) and can be seen as 
a sequence of $small$ steps, execution from top-down, support no correct and 
support no parallel codes.    These leads to non-parallel,
efficient yet messy and incorrect code.

(4) is  simply a high-level approach without any use of  lemmas and can be seen as an empty 
 sequence.
We consider this the worst, leading to clean yet inefficient code.

\section{Conclusion}

Note that  we introduce an  interesting kind of agents called class agents.
 A class agent is a collection of agents in a compressed form.

Our ultimate goal is to implement the computability
logic web\cite{Jap03,Jap08}  which is a promising approach to
reaching general AI.  New ideas in this paper  will be useful for organizing agents and
their knowledgebases in \colw.

We now discuss how the machine executes \colw. An important point is that it must first check the $validity$ of \colw\ given by the programmer.
This is not an easy task which may require new, sophisticated inductions. In this sense, it is a  
$proof$-$carrying$ code.

\bibliographystyle{ieicetr}

\end{document}